\documentclass[10pt,twocolumn,letterpaper]{article}

\usepackage{subcaption}
\usepackage{pifont} 
\usepackage{tabularx} 
\usepackage{algorithm, algpseudocode} 
\usepackage{fontawesome} 
\usepackage[dvipsnames]{xcolor} 
\usepackage{booktabs} 

\usepackage{soul}
\usepackage{iccv}
\usepackage{times}
\usepackage{epsfig}
\usepackage{graphicx}
\usepackage{amsmath}
\usepackage{amssymb}
\usepackage{caption}
\newcolumntype{Y}{>{\centering\arraybackslash}X} 
\algnewcommand{\Inputs}[1]{%
  \State \textbf{Inputs:}
  \Statex \hspace*{\algorithmicindent}\parbox[t]{.8\linewidth}{\raggedright #1}
}
\algnewcommand{\Outputs}[1]{%
  \State \textbf{Outputs:}
  \Statex \hspace*{\algorithmicindent}\parbox[t]{.8\linewidth}{\raggedright #1}
}

\usepackage[pagebackref=true,breaklinks=true,letterpaper=true,colorlinks,bookmarks=false]{hyperref}

\iccvfinalcopy 

\def\iccvPaperID{18} 

\ificcvfinal\fi

\begin{document}

\title{Spatio-Temporal Analysis of Patient-Derived Organoid Videos Using Deep Learning for the Prediction of Drug Efficacy}


\author{Leo Fillioux\textsuperscript{1}, Emilie Gontran\textsuperscript{2}, Jérôme Cartry\textsuperscript{2}, Jacques RR Mathieu\textsuperscript{2}, Sabrina Bedja\textsuperscript{2},\\Alice Boilève\textsuperscript{2}, Paul-Henry Cournède\textsuperscript{1}, Fanny Jaulin\textsuperscript{2}, Stergios Christodoulidis\textsuperscript{1}, Maria Vakalopoulou\textsuperscript{1}
\\ \normalfont
\normalsize\textsuperscript{1}CentraleSupélec, Université Paris Saclay\quad
{\tt\small first.last@centralesupelec.fr}
\\
\normalsize\textsuperscript{$2$}Gustave Roussy\quad{\tt\small first.last@gustaveroussy.fr}
}

\maketitle
\ificcvfinal\thispagestyle{empty}\fi

\begin{abstract}
Over the last ten years, Patient-Derived Organoids (PDOs) emerged as the most reliable technology to generate ex-vivo tumor avatars. PDOs retain the main characteristics of their original tumor, making them a system of choice for pre-clinical and clinical studies. In particular, PDOs are attracting interest in the field of Functional Precision Medicine (FPM), which is based upon an ex-vivo drug test in which living tumor cells (such as PDOs) from a specific patient are exposed to a panel of anti-cancer drugs. 
Currently, the Adenosine Triphosphate (ATP) based cell viability assay is the gold standard test to assess the sensitivity of PDOs to drugs. The readout is measured at the end of the assay from a global PDO population and therefore does not capture single PDO responses and does not provide time resolution of drug effect. To this end, in this study, we explore for the first time the use of powerful large foundation models for the automatic processing of PDO data. In particular, we propose a novel imaging-based high-throughput screening method to assess real-time drug efficacy from a time-lapse microscopy video of PDOs. The recently proposed SAM algorithm for segmentation and DINOv2 model are adapted in a comprehensive pipeline for processing PDO microscopy frames. Moreover, an attention mechanism is proposed for fusing temporal and spatial features in a multiple instance learning setting to predict ATP. We report better results than other non-time-resolved methods, indicating that the temporality of data is an important factor for the prediction of ATP. Extensive ablations shed light on optimizing the experimental setting and automating the prediction both in real-time and for forecasting.

\end{abstract}


\section{Introduction}
Precision medicine aims to optimize the choice of drug given the characteristics of the patient, so as to optimize certain aspects such as the efficacy of the treatment or quality of life of the patient. Although doing this on a case-by-case basis by a clinician seems impractical, artificial intelligence-driven tools help guide this approach. In this objective, FPM \cite{letai2017functional} bases this optimization on tests performed on live patient cells.

Patient-derived organoids (PDOs) have gained great interest over the last few years as they represent minimalistic models to mimic essential features from the tissue they originate from.
In the context of cancer therapy, drug efficiency can be limited due to the development of resistance in patients as well as other evolutionary changes in the tumors over time.
PDOs represent a good testbed for physicians, researchers, and patients to assess personalized drug efficacy, based on tumor-specific patient characteristics.

The gold standard method to assess drug efficacy on cells relies on Adenosine Triphosphate (ATP), an energy molecule released in active cells through metabolic reactions. ATP is a biomarker for evaluating the number of viable cells. The quantity of released ATP measured by luminescence is proportional to the number of living cells. Thus, ATP quantity assessed by luminescence counts serves as a readout to evaluate drug efficacy with the estimation of remaining living cells in the experimental sample.
This test is easily implementable in bench labs and reliable for assessing the global cell population response to drug exposure \cite{phan2019simple}. However, since it causes cell lysis, the ATP test is a destructive assay, that does not allow to assess real-time organoid drug response, and post-ATP observations to evaluate long-term viability changes and drug resistance.

The emergence of large foundation models in various fields of machine learning has allowed for novel solutions in a variety of downstream tasks. Notably, SAM (Segment Anything Model) \cite{kirillov2023segany} or SEEM (Segment Everything Everywhere All at Once) \cite{zou2023segment} have facilitated segmentation by providing a general model for segmentation from various types of prompts, which is extremely valuable for tasks for which little to no annotations can be found. Similarly, self-supervised \cite{chen2020simple, bardes2022vicreg, goyal2021selfsupervised, caron2021emerging} or unsupervised models such as DINOv2 \cite{oquab2023dinov2} provide high-quality descriptors for data and facilitate the training for downstream tasks with relatively small data and task-specific samples. This training setting allows the extracted features to be task-agnostic and, therefore, to adapt more easily to new tasks.
Building on these recent advances, in this paper, we propose a new high-throughput screening method for the analysis of PDOs testing multiple drugs and high-quality videos. Indeed, processing of this data is usually performed in a manual and time-consuming setting.

In this paper, our contributions are the following.
To our knowledge, we present the first fully automatic method for processing high-quality videos of PDOs, conducting a spatio-temporal analysis for the prediction of ATP.
We propose an efficient, automatic, and accurate prompt engineering paradigm taking into account the temporal characteristics of PDOs for using SAM without the need for additional training.
Finally, we explore powerful recent foundation models for the spatio-temporal representation of PDOs for the first time coupling them with time sequence modeling and multiple instance learning.

An extensive experimental analysis was performed to identify the best representations for PDOs as well as the most informative time frames, which we used for the accurate prediction of ATP. This opens the possibility to integrate any other clinical endpoint, to match PDO drug sensitivity and patient clinical response.



\section{Related work}

\textbf{Foundation Models in Computer Vision.}
Foundation models are precious resources and stimulate research by both providing a strong solution for a specific task and by proposing a novel innovative approach. The authors of the ``Florence'' model \cite{yuan2021florence} introduced a method for learning joint visual-textual features that can be adapted to a multitude of joint tasks. The CLIP model \cite{radford2021learning} uses a contrastive learning approach in order to learn image embeddings, guided by an associated textual description of the image, which can transfer zero-shot to a wide variety of tasks. This approach has led to many extensions \cite{li2022grounded, mu2021slip, li2022blip}. DINOv2 \cite{oquab2023dinov2} is a deep learning model trained in a self-supervised manner that leverages more stable training at a bigger scale for better taking advantage of large datasets. The model is trained on a dataset of 142M images. It is based on the ViT architecture \cite{dosovitskiy2021image} and is made available in different sizes. Very recently, foundation models for image segmentation such as SegGPT \cite{wang2023seggpt}, SEEM \cite{zou2023segment}, and SAM \cite{kirillov2023segany} have surfaced. SAM provides a strong zero-shot transfer approach by combining a powerful image encoder with a prompt encoder which can adapt to multimodal prompts. Its convincing adaptability led to many adaptations for videos \cite{yang2023track}, lightweight versions \cite{zhao2023fast}, or medical applications \cite{wu2023medicalsam, liu2023samm}. While these models show impressive results and generalizability, their adaptations to medical applications often show limitations. For example, recent generative models such as GLIDE \cite{nichol2022glide} or DALL-E \cite{ramesh2021zeroshot} show impressive results on natural images but struggle to generate realistic medical images without prior fine-tuning \cite{kather2022medical, adams2022does}.


\textbf{Time Sequence Modeling.}
Modeling the temporal aspect of data can be achieved either on extracted features of the frames or directly in an end-to-end fashion on the videos. Once features have been extracted from frames of the videos, standard sequence models such as LSTMs \cite{hochreiter1997long} or Transformers \cite{vaswani2017attention, zerveas2020transformerbased} can be applied for a variety of tasks. They aim to capture dependencies between elements of a sequence, either in the form of recurrent neural networks or self-attention mechanisms. State space modeling is gaining attraction in sequence modeling, especially for long sequences \cite{gu2022efficiently}, and more specifically for modeling time series \cite{zhang2023effectively}. For end-to-end video modeling, Transformer adaptations are very popular such as the Video Swin Transformer \cite{liu2021video}, ViViT \cite{arnab2021vivit}, or the TimeSformer \cite{bertasius2021spacetime}. We are also beginning to see adaptations of large language models (LLMs) for video understanding tasks \cite{chen2023videollm}. However, even if these models are available, their application to PDOs has not yet been explored and investigated, especially in a low-availability data regime. 

\textbf{Analysis of PDO Microscopy Images.} With the emergence of PDOs as a key technology for FPM in the last few years, machine learning methods to analyze this data have followed. Earlier studies \cite{bian2021tracking, Matthews2022OrganoID} looked for methods for tracking and analyzing the dynamics of PDOs through time. Authors of "D-CryptO" \cite{Abdul2022DCryptO} proposed to classify types of organoids based on their morphology. Later research sought to use organoid analysis pipelines for other downstream tasks, such as the prediction of kidney differentiation \cite{Park2023Kidney} or the prediction of a biomarker of Huntington's disease \cite{Metzger2022Huntington}. Recently \cite{bian2023insatp} have developed a pipeline for predicting ATP from microscopy images at single time points in a multiple instance learning setting. However, all these methods focus on analyzing single representations of organoids, losing a lot of information that arises from their temporal dynamics for different drug treatments. Moreover, they are mainly based on classical processing techniques, such as the extraction of predefined visual features, the extraction of ResNet features, or segmentation through biological staining.

\section{Methods}
\begin{figure*}[t]
\begin{center}
\includegraphics[width=0.95\linewidth]{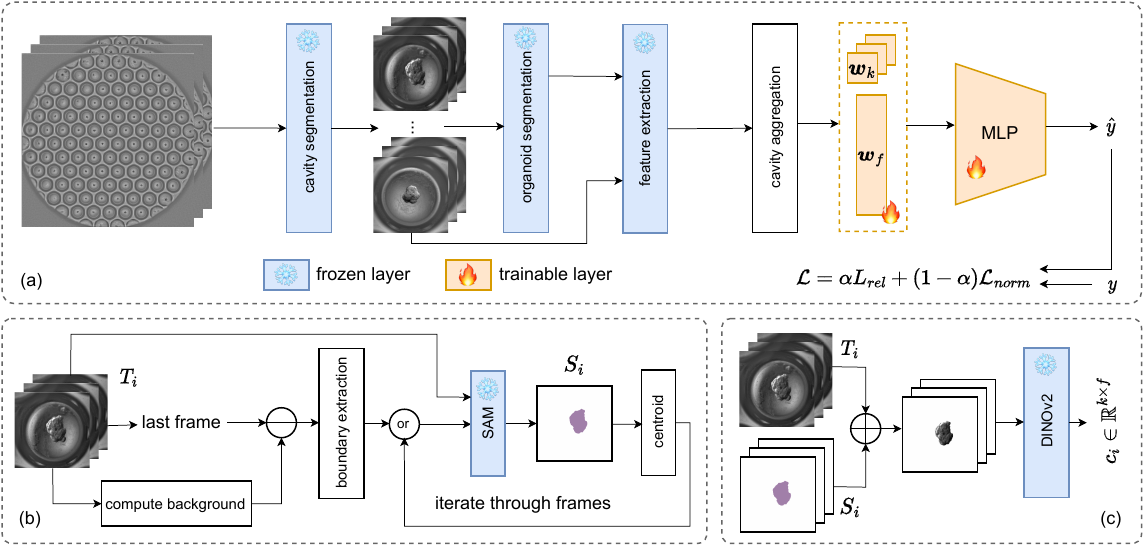}
\end{center}
   \caption{Illustration of our proposed pipeline. (a) The overall pipeline taking as input the whole well timelapse and outputting the predicted ATP $\hat{y}$. (b) Organoid segmentation based on automatic generation of prompts for SAM \cite{kirillov2023segany}. (c) Feature extraction given input frames and associated segmentation maps focusing in the region of interest and using DINOv2 model \cite{oquab2023dinov2}.}
\label{fig:method_pipeline}
\end{figure*}

In this section, we will start by describing the specificities of the dataset and the preprocessing steps, and then proceed to an in-depth presentation of the proposed method. Specifically, we will present the methodology for the segmentation of the cavities and organoids by employing SAM \cite{kirillov2023segany}, then we will present the details of the feature extraction from the organoid regions using DINOv2 \cite{oquab2023dinov2}. The model for the final prediction of the ATP given the features from the cavities under the multiple instance learning framework will then be presented. Figure \ref{fig:method_pipeline} illustrates the proposed pipeline.

\subsection{Data}

To capture real-time information about drug efficacy on patient organoids, bright-field imaging techniques represent a good non-invasive alternative to the ATP bioluminescence test. However, having single-organoid resolution may be a challenging task when working with the standard experimental setup where organoids are embedded in hydrogel droplets, which results in heterogenous organoid distribution at different depths and subsequently heterogeneous organoid sizes, making the drug efficacy assessment less accurate. To overcome these issues, we developed a high-throughput experimental and imaging pipeline. It consists in using high-throughput cell culture systems with 96-well plates containing 500 $\mu$m diameter cavities. Each cavity contains one organoid, while the setting allows for the parallel follow-up of individual organoids with a homogeneous depth distribution. Testing multiple drugs in parallel thus becomes easier with one image containing the information of multiple organoids under the same treatment. One type of drug at a given concentration is applied to each well, and the ATP can only be measured on the well level. Figure \ref{fig:methods_org_preprocessing} shows one well containing multiple cavities together with their automatic segmentation. A total of 116 wells are imaged, from which we detect 8241 cavities, and therefore 8241 organoids.

A cavity is imaged in a set of frames $T = \{T_1, ..., T_f\}, T_i\in\mathbb R^{w\times h}$, where $w$ and $h$ are the frame dimensions, and $f$ the number of frames in each timelapse. In this study, we developed a novel pipeline to segment each cavity
$S= \{S_1, ..., S_f\}, S_i\in\mathbb R^{w\times h}$ and using $S$ and $T$, we extract interesting features from each organoid.
Finally, we model a well by a set of cavity features $\mathcal C = \{c_1, \ldots, c_n\} \in \mathbb R^{n\times k\times f}$, where $n$ is the number of cavities in the well, and $k$ the dimensionality of the feature space. The final ATP value is defined as $y \in \mathbb R$.

The patient-derived organoids (from now on called organoids, for simplicity) are extracted from colorectal cancer patients. 
In this analysis, we do not consider the concentration of the drugs and only observe their effects indirectly with the ATP. Drugs are introduced one day after the organoids have been formed in the cavities. The wells are imaged every 30 minutes for 100 hours, resulting in $f=200$ frames for each well. Wells are digitized using an Agilent Lionheart FX digital microscope. Note that some experiments could take slightly longer than 200 frames, for those cases, only the last 200 frames of the experiments were retained, in order to keep the measurement of the ATP at the last frame.

\subsection{Preprocessing.}
Each scan consists of acquiring quarters of wells at three different depth levels, resulting in 12 images for each time point. A z-projection based on Sobel filters \cite{kanopoulos1988design} is used to project the images into the same plane, before stitching all four corners into a single image. Artifacts appear through time (e.g. evaporation of the liquid impacting the contrast), which may impact the segmentation pipeline. To account for this, we normalize the contrast of the timelapse through time. Finally, all frames of a timelapse are then co-registered using SIFT-based features \cite{Lowe2004}. All preprocessing is performed using ImageJ \cite{schneider2012image}. 
Figures \ref{subfig:methods_pre_all}-\ref{subfig:methods_pre_norm} show all steps of the preprocessing from the 12 input images per frame to the normalized frame.

\subsection{Automatic Segmentation of Regions of Interest}

In this study, we propose and develop an automatic pipeline based on SAM \cite{kirillov2023segany} for the automatic segmentation of regions of interest in the video of the organoids.
For accurate segmentation, various characteristics of these regions are used and different prompt engineering strategies have been developed to handle the nature of the data and their temporal information.

\begin{figure}[b]
\begin{center}
\begin{subfigure}{.29\linewidth}
    \includegraphics[width=\linewidth]{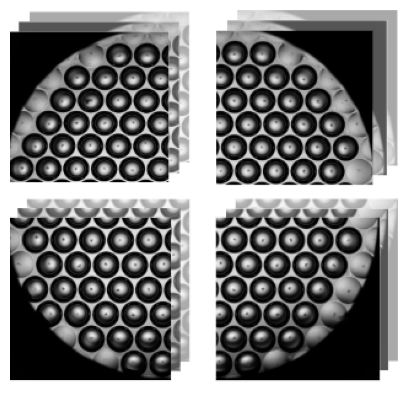}
    \caption{}\label{subfig:methods_pre_all}
\end{subfigure}
\hfill\begin{subfigure}{.29\linewidth}
    \includegraphics[width=\linewidth]{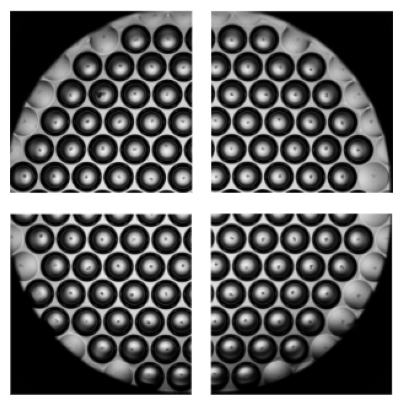}
    \caption{}\label{subfig:methods_pre_proj}
\end{subfigure}
\hfill
\begin{subfigure}{.29\linewidth}
    \includegraphics[width=\linewidth]{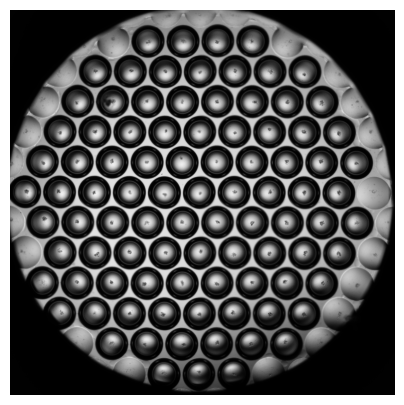}
    \caption{}\label{subfig:methods_pre_stitched}
\end{subfigure}
\bigskip
\begin{subfigure}{.29\linewidth}
    \includegraphics[width=\linewidth]{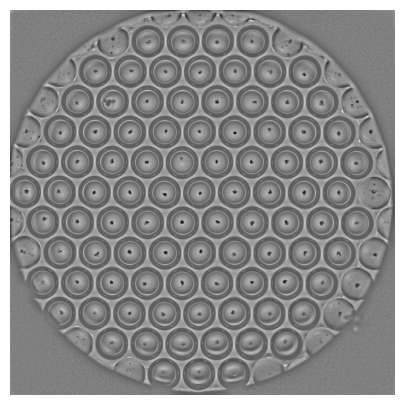}
    \caption{}\label{subfig:methods_pre_norm}
\end{subfigure}
\begin{subfigure}{.29\linewidth}
    \includegraphics[width=\linewidth]{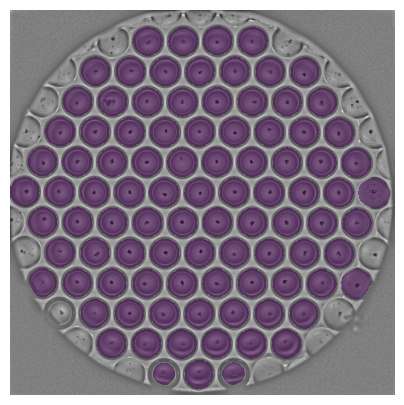}
    \caption{}\label{subfig:methods_pre_mask}
\end{subfigure}
\vspace{-2\baselineskip}

\end{center}
   \caption{Example of the steps needed for preprocessing and cavity segmentation. (a) All images needed to construct one frame (4 corners at 3 different z-levels each). (b) Projected corners. (c) Stitched frame. (d) Local contrast normalization applied to the stitched frame. (e) Cavity segmentation masks.}
\label{fig:methods_org_preprocessing}
\vspace{-\baselineskip}
\end{figure}

\textbf{Cavity Segmentation.}
The objective is to segment the individual cavities contained in the well, in order to represent the well as a set of cavities. We assume that the detection of the cavities only needs to be run in a single frame, as the frames of the video have been co-registered. The well is considered as the mask with the biggest surface area containing the center pixel in the frame is considered, as it is the biggest object in the frame. Given the candidate masks $\mathcal M = \{m_1, \ldots, m_n\}$, $m_{\text{well}}=m_{\textit{argmax}_i\{\text{area}(m_i)\}}$. The cavities are defined as other regions detected by SAM, which have a circularity above a certain threshold and which are included in the detected well. For each point $b_i$ on the boundary $B$ of a detected region $m$, we define $\mathcal P = \{\max_i(\text{dist}(b_i, B))\}$, which represents for each point on the contour, the distance to the point on the contour that is furthest away. The ratio $\frac{\min \mathcal P}{\max \mathcal P}$ defines the circularity. On a perfect circle, all values in $\mathcal P$ should be equal to the circle's diameter and therefore have $\text{circularity}=1$.
Once the cavities have been detected on a single frame, a crop around the cavity is extracted for all frames, giving a few dozen cavity timelapses per well. Figure \ref{subfig:methods_pre_mask} shows an example of the cavity segmentation masks on a normalized frame.

\begin{figure}[b]
\begin{center}
\begin{subfigure}{.3\linewidth}
    \includegraphics[width=\linewidth]{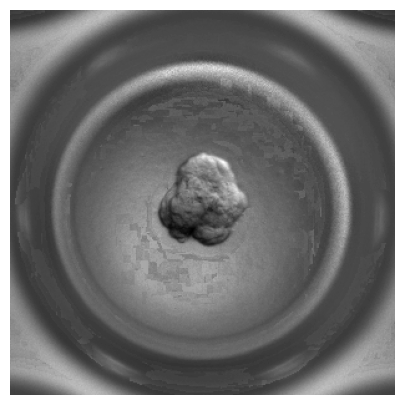}
    \caption{}\label{subfig:methods_org_last}
\end{subfigure}
\hfill\begin{subfigure}{.3\linewidth}
    \includegraphics[width=\linewidth]{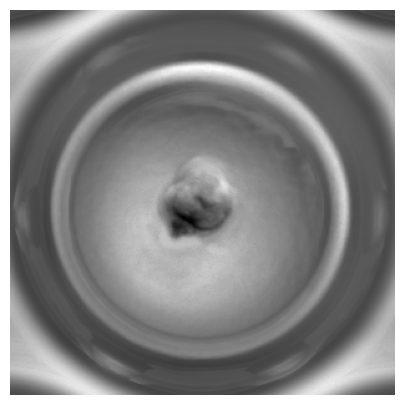}
    \caption{}\label{subfig:methods_org_mean}
\end{subfigure}
\hfill
\begin{subfigure}{.3\linewidth}
    \includegraphics[width=\linewidth]{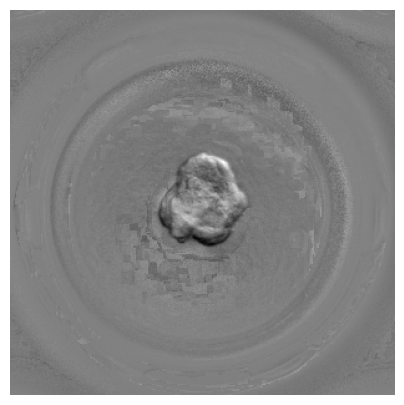}
    \caption{}\label{subfig:methods_org_diff}
\end{subfigure}
\bigskip
\begin{subfigure}{.3\linewidth}
    \includegraphics[width=\linewidth]{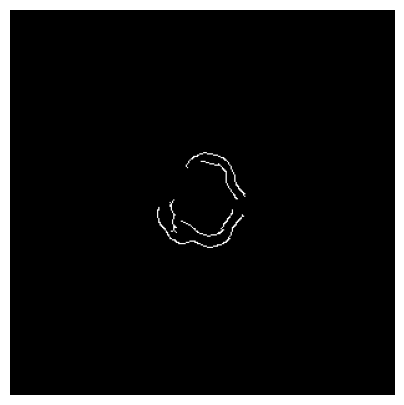}
    \caption{}\label{subfig:methods_org_borders}
\end{subfigure}
\hfill\begin{subfigure}{.3\linewidth}
    \includegraphics[width=\linewidth]{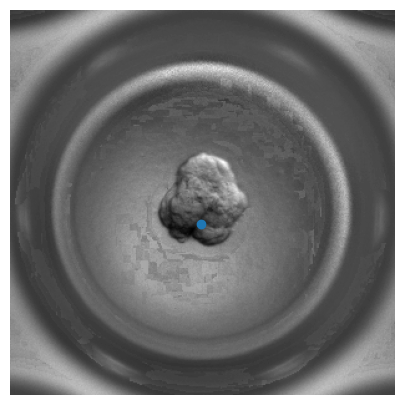}
    \caption{}\label{subfig:methods_org_prompt}
\end{subfigure}
\hfill
\begin{subfigure}{.3\linewidth}
    \includegraphics[width=\linewidth]{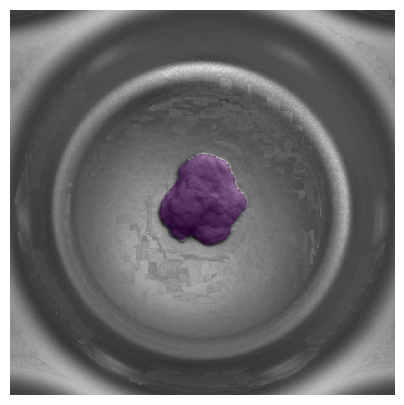}
    \caption{}\label{subfig:methods_org_mask}
\end{subfigure}
\vspace{-2\baselineskip}

\end{center}
   \caption{Example of the steps used in the generation of the prompt for the segmentation of the last frame of a cavity timelapse. (a) Last frame of the timelapse. (b) Mean frame across time. (c) Difference between the last frame and the mean frame. (d) Result of Canny edge detection on (c). (e) Generated prompt for the frame. (f) Predicted mask.}
\label{fig:methods_org_seg}
\vspace{-\baselineskip}
\end{figure}

\textbf{Organoid Segmentation.} \label{methods_org_seg}
Segmentation of organoids is performed on the extracted videos of the cavities by designing the proper prompts for SAM \cite{kirillov2023segany}. The first prompt is generated for the last frame of the timelapse $T_f$ (as it is the frame where the organoid is the largest, and it is, therefore, more likely to find a point inside the region of interest). Canny edge detection \cite{canny1986computational} is used to extract rough contours from the last frame, after having subtracted the average frame over time, in order to remove the background. These rough contours are then filtered using a series of binary morphological operators, before extracting a centroid, which is used as a positive point prompt for SAM. Figure \ref{fig:methods_org_seg} shows an example of these steps for the last frame.
Once the mask for the last frame has been generated, the prompts will be generated backward from the last frame to the first frame. The center of the organoid mask is extracted from the prediction and an exponentially weighted average from the prompts from frames $T_{t+1}$ to $T_{t+10}$ is used to generate the prompt for frame $T_t$. Multimask output is used to generate the top 3 masks, and the mask which has the highest Dice score with the generated mask from the previous frame ($T_{t+1}$) is chosen. Postprocessing is performed on the mask by only selecting the detected region with the highest surface area. Algorithm \ref{alg:org_seg} represents the whole organoid segmentation pipeline.

\begin{algorithm}[t]
    \caption{Organoid segmentation.}
  \begin{algorithmic}
    \Inputs{frames $T = \{T_1, ..., T_f\}, T_i\in\mathbb R^{w\times h}$}
    \Outputs{segmentations $S= \{S_1, ..., S_f\}, S_i\in\mathbb R^{w\times h}$}
    
    \State estimated\_boundary$ = \textit{Canny}(T_f - \textit{mean}(T))$
    \State prompt\_f = \textit{centroid}(estimated\_boundary)
    \State $S_f = \textit{SAM}(\text{prompt\_f}, T_f)$
    \State all\_prompts = \{prompt\_f\}
    \For{i=(f-1); i=1; i - -}
      \State prompt\_i = \textit{exp\_weighted\_average}(all\_prompts, 10)
      \State $S_i = \textit{SAM}(\text{prompt\_i}, T_i)$
      \State all\_prompts.insert(prompt\_i)
    \EndFor
  \end{algorithmic}
\label{alg:org_seg}
\end{algorithm}
\vspace{-\baselineskip}


\subsection{Feature Extraction} \label{methods_features}
Features are extracted from each frame of the cavity using the DINOv2 \cite{oquab2023dinov2} model, which produces task-agnostic visual features from images. A lighter version (86M parameters) of the model is used, which produces $k$-dimensional feature vectors. Features are extracted per frame, using a crop of the cavity around the organoid, and by masking everything outside of the regions of interest, giving $c_i=\text{DINOv2}(T_i \odot S_i)$. The model generates feature vectors of dimension $k=768$ for each time frame. 


\subsection{ATP Prediction} \label{methods_atp_pred}
Having access to features for each individual cavity but only having the ATP measure on the well level, the multiple instance learning setting seems to be the best approach to the problem. Each set of cavities $\mathcal C$ is associated with an ATP measure $y$. The ATP prediction model $\mathcal M$ should map an input $\mathcal C \in \mathbb R^{n\times k\times f}$ to an output $\hat{y} \in \mathbb R$.
The aggregation of the cavity representations on the well level is done by mean pooling in order to obtain a well-level representation $\textit{mean}(\mathcal C)\in \mathbb R^{k\times f}$.

Each time frame may have a different impact on the prediction of the ATP. In fact, we expect later time frames to have more impact than earlier time frames. We give the model the freedom to learn how much weight to put to each time frame using a normalized weight vector $\boldsymbol w_t \in \mathbb R^f$ that is used in a weighted average across time frames. This ensures that all cavities across all wells share the same importance for a given time frame.
A feature-wise weight vector $\boldsymbol w_k \in \mathbb R^k$ is used in order to learn the relative importance of each feature that is shared across all wells.

The last element of the model is a multilayer perceptron (MLP) which is simply composed of four linear layers, with PReLU \cite{he2015prelu} activation function.

\begin{align}
\begin{split}
    \mathcal M \colon \mathbb R^{n\times k\times f} &\rightarrow \mathbb R \\
    \mathcal C &\mapsto \textit{MLP}(\textit{mean}(\mathcal C) \cdot \frac{\boldsymbol w_t}{\Vert \boldsymbol w_t \Vert} \odot \boldsymbol w_k)
\end{split}
\end{align}

The loss function used for training is composed of two elements, a relative $L_1$ loss, to directly optimize for the  mean absolute percentage error (MAPE), and a $L_1$ loss, which has been normalized with the maximum ATP value in the training set to have a comparable range to the relative $L_1$ loss.

\begin{align}
\mathcal L_{\textit{rel}}(y, \hat{y}) &= \frac{|y-\hat{y}|}{y} \\
\mathcal L_{\textit{norm}}(y, \hat{y}, y_{\textit{max}}) &= \frac{|y-\hat{y}|}{y_{\textit{max}}} \\
\mathcal L(y, \hat{y}, y_{\textit{max}}) &= \alpha \mathcal L_{\textit{rel}} + (1-\alpha) \mathcal L_{\textit{norm}}
\end{align}

Where $y$ is the ATP ground truth, $\hat{y}$ the output of our model, $y_{\textit{max}}$ the maximum value of ATP in the training set, and $\alpha$ the weight given to $\mathcal L_{\textit{rel}}$. A relative $L_1$ loss is used because of the varying range that the ATP can take within experiments. The normalized $L_1$ loss is used to ensure that the model still has an acceptable performance on higher values of ATPs.


\subsection{Implementation Details} \label{methods_implementation}

Over all experiments, training was performed under the same scheme. Cross-validation is performed over 4 splits, on the well level, resulting in training sets of size 87 and validation sets of size 29. The reported results are the average over all folds of the validation sets.
All trainings were performed using the PyTorch deep learning library \cite{paszke2019pytorch} in Python. The AdamW \cite{loshchilov2019adamw} optimizer was used with a learning rate of $1\cdot10^{-3}$, weight decay of $1\cdot10^{-1}$, and a batch size of 8. The model was trained over 2000 epochs, and early stopping with a patience of 200 epochs was used. The $\alpha$ parameter which balances both terms of the loss was set at $\alpha=0.5$ after grid search. All trainings were accelerated using Nvidia Tesla V100 GPUs.

\section{Experiments}


\begin{table*}
\begin{center}
\resizebox{0.75\linewidth}{!}{
\begin{tabular}{c c c c c c c c c}
\toprule
\multicolumn{1}{c}{ATP bin ($10^5$)} & \multicolumn{1}{c}{${[}1.1, 2.8)$} & \multicolumn{1}{c}{${[}2.8, 4.5)$} & \multicolumn{1}{c}{${[}4.5, 6.2)$} & \multicolumn{1}{c}{${[}6.2, 8.0)$} & \multicolumn{1}{c}{${[}8.0, 9.7)$} & \multicolumn{1}{c}{${[}9.7, 11.4)$} & \multicolumn{1}{c}{${[}11.4, 13.1)$} & \multicolumn{1}{c}{${[}13.1, 14.9{]}$}  \\
\midrule
Count & 23 & 19 & 11 & 15 & 0 & 7 & 17 & 24 \\
MAPE $\downarrow$
& 0.34 & 0.15 & 0.08 & 0.07 & NA & 0.10 & 0.23 & 0.12 \\
\bottomrule
\end{tabular}}
\end{center}
\caption{Distribution of ATP values in our dataset, with associated MAPE.}
\label{tab:atp_histogram}
\end{table*}

To evaluate the performance of the method we used the MAPE and Pearson correlation coefficient metrics. All performances are measured on each validation set, utilizing all four models trained on the corresponding training sets.
The MAPE, measures the mean relative error between the prediction $\hat{y}$ and the label $y$. Because of the very wide range of values of the ATP (both within an experiment and between different experiments) as presented in Table~\ref{tab:atp_histogram}, the MAPE appears suitable for ensuring comparability, in particular for future studies which may use datasets with different distributions of ATP. Lower values of MAPE indicate better performance.

\begin{equation}
    \textit{MAPE}(y, \hat{y}) = \frac{\vert y-\hat{y}\vert }{y}
\end{equation}

The Pearson correlation coefficient quantifies the linear correlation between the set of predictions $\hat{Y}$ and the set of labels $Y$. Higher values of the Pearson correlation coefficient indicate better performance.

\begin{equation}
    \textit{Pearson}(Y, \hat{Y}) = \frac{\sum_i(Y_i-\bar{Y})(\hat{Y}_i-\bar{\hat{Y}})}{\sqrt{\sum_i(Y_i-\bar{Y})^2(\hat{Y}_i-\bar{\hat{Y}})^2}}
\end{equation}

\begin{figure}[t]
\begin{center}
\begin{subfigure}{.27\linewidth}
    \includegraphics[width=\linewidth]{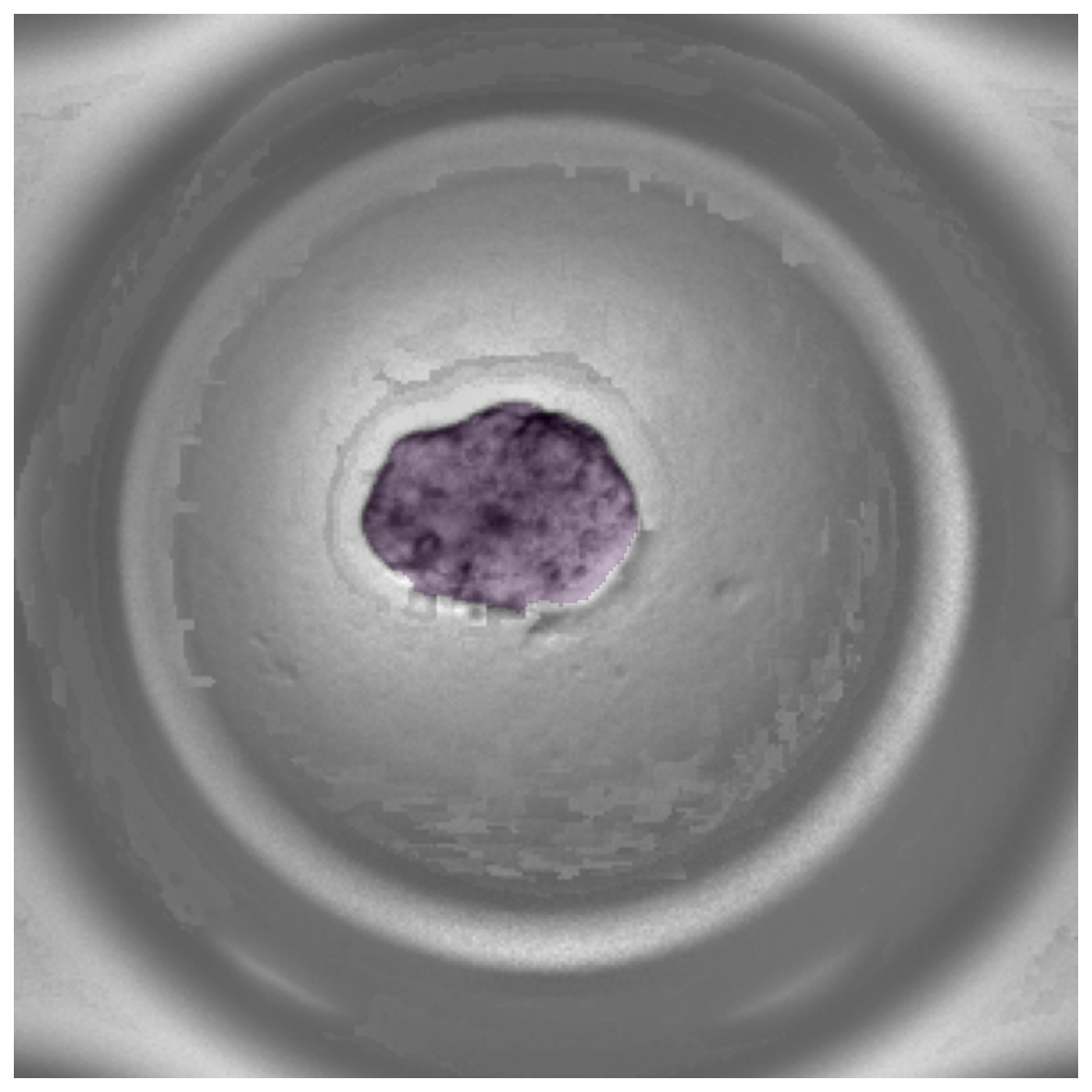}
    \caption{}\label{subfig:seg_ours}
\end{subfigure}
\hfill\begin{subfigure}{.27\linewidth}
    \includegraphics[width=\linewidth]{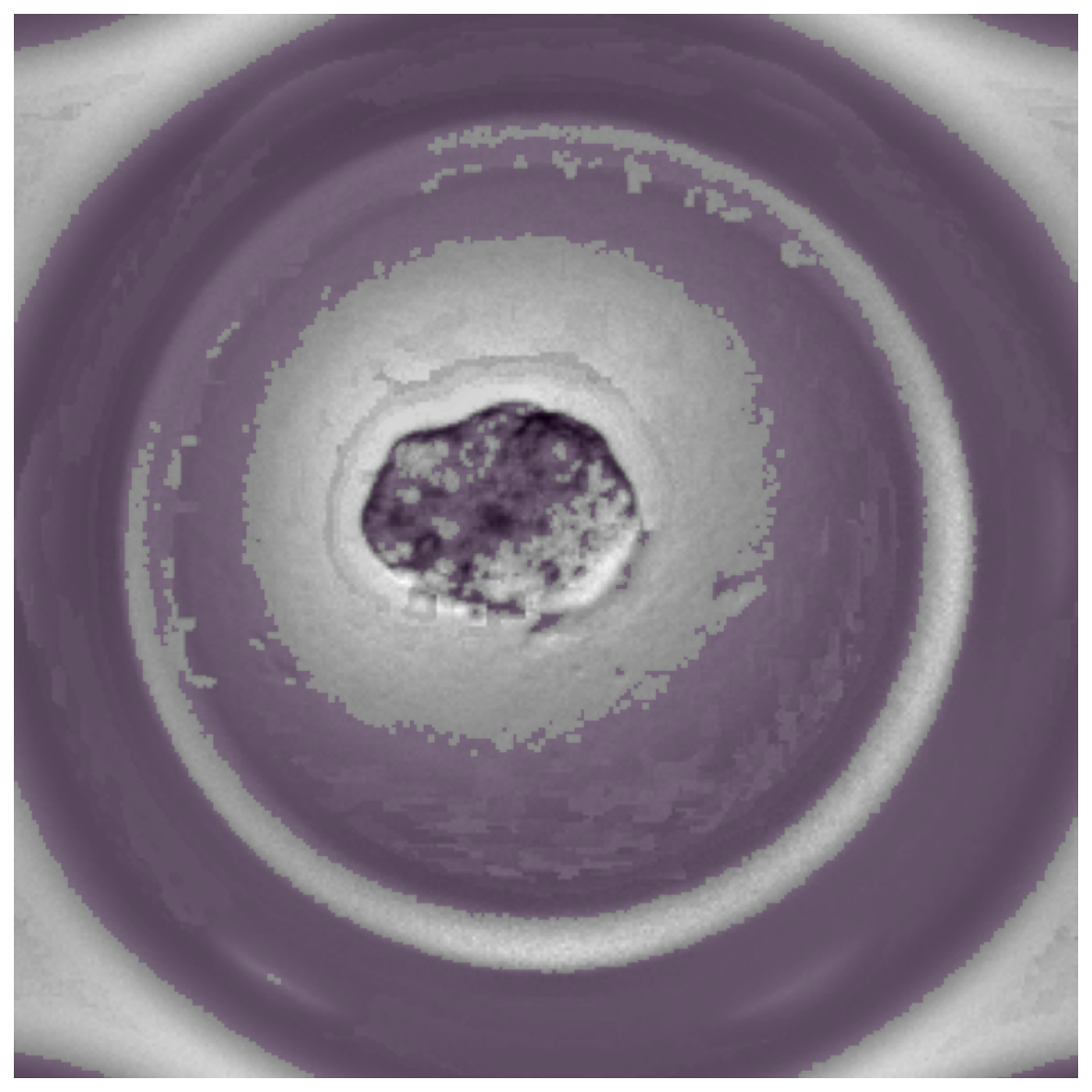}
    \caption{}\label{subfig:seg_th}
\end{subfigure}
\hfill
\begin{subfigure}{.27\linewidth}
    \includegraphics[width=\linewidth]{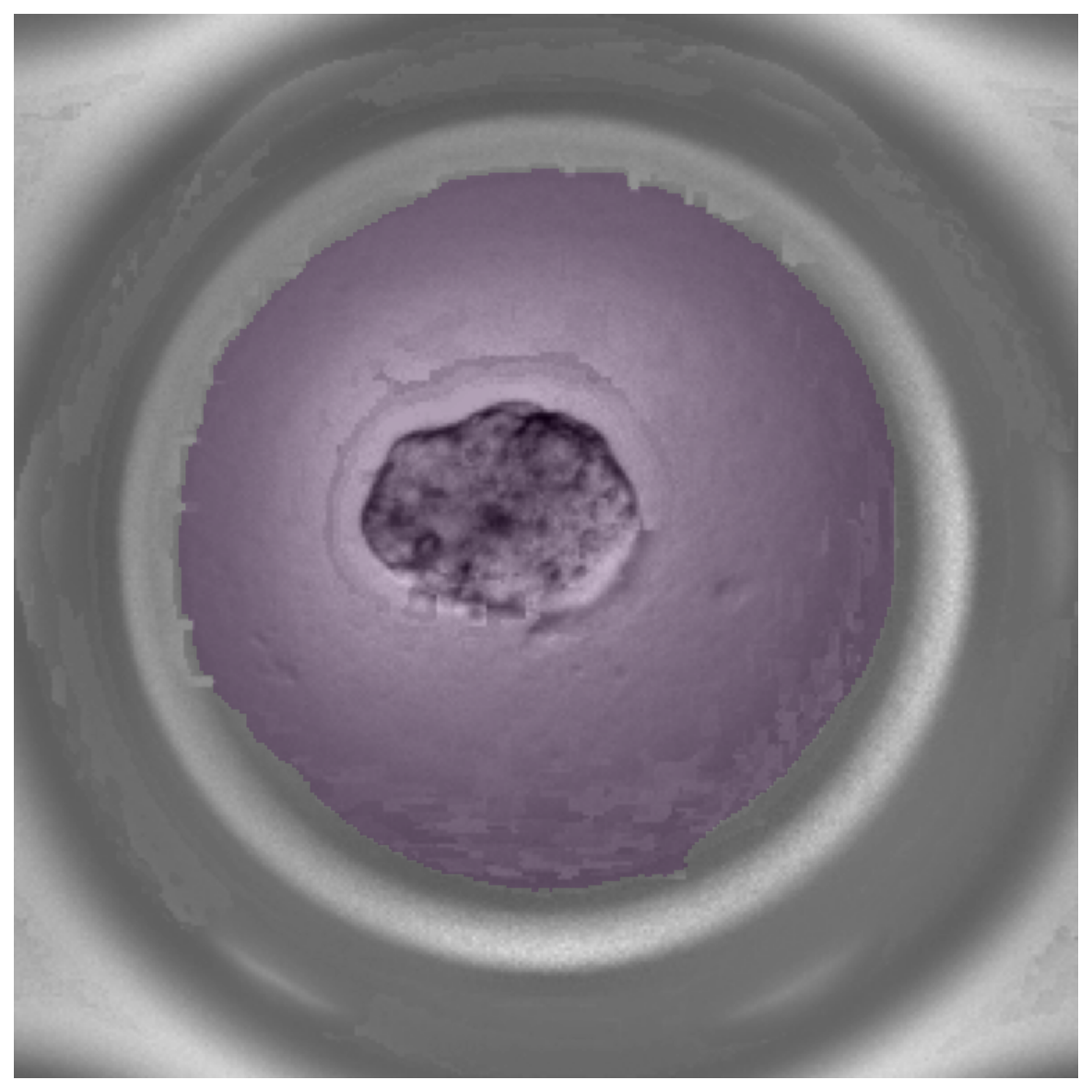}
    \caption{}\label{subfig:seg_sam}
\end{subfigure}
\vspace{-1\baselineskip}

\end{center}
   \caption{Comparison of three methods of segmentation. (a) Our proposed method. (b) Otsu thresholding
   based segmentation. (c) Using SAM \cite{kirillov2023segany} without any prompts and choosing the mask with the highest predicted IOU.}
\label{fig:segmentation_methods}
\vspace{-\baselineskip}
\end{figure}

Figure \ref{fig:segmentation_methods} qualitatively compares three methods of organoid segmentation to justify the superiority of our proposed method over simple intensity-based methods (Otsu thresholding) and justifies the use of our prompt generation for the use of SAM \cite{kirillov2023segany}.

\subsection{Results}
With our proposed high-throughput screening method, we obtain an average MAPE and Pearson correlation coefficient on the four validation sets of $0.1755$ and $0.9214$, respectively. The relative error of $17.55\%$ is to be put in context with the wide range of ATP values in our dataset, with a ratio between the highest and lowest values of ATP of 14. Table \ref{tab:atp_histogram} shows the distribution of ATP values along with the corresponding performance of the model in terms of MAPE in each bin.  The model performs worst on the bin with the smallest values of ATP, which may be explained by the precision of the measurement of ATP.

\begin{table}[b]
\begin{center}
\resizebox{\columnwidth}{!}{
\begin{tabular}{c c c c c }
\toprule
\multicolumn{1}{c}{Features} & \multicolumn{1}{c}{Classical} & \multicolumn{1}{c}{ResNet50~\cite{he2015resnet}} & \multicolumn{1}{c}{VICReg~\cite{bardes2022vicreg}} & \multicolumn{1}{c}{DINOv2~\cite{oquab2023dinov2}}  \\
\midrule
MAPE $\downarrow$
& 0.3173          & 0.2174       
& \underline{0.1847}  & \textbf{0.1755} \\
Pearson $\uparrow$
& 0.7883 & \underline{0.8960}
& 0.8939  & \textbf{0.9214} \\
\bottomrule
\end{tabular}}
\end{center}
\caption{Comparison of MAPE and Pearson correlation coefficient for four methods of feature extraction. Best results are indicated in \textbf{bold}, and second best results are \underline{underlined}.
}
\label{tab:feature_ablation}
\end{table}

\textbf{Feature Extraction.}
In Table \ref{tab:feature_ablation}, we explore the impact of four different feature extraction methods on the final results. Different classical and deep learning features have been explored to highlight the best representations for organoids. The pyradiomics Python package \cite{vangriethuysen2017computational} allows the extraction of imaging features from regions of interest. For this study, we extracted features related to first-order statistics, gray-level matrices, and shape features. A total of $93$ features have been extracted per frame.
ResNet50 \cite{he2015resnet} (pre-trained on ImageNet \cite{deng2009imagenet}) is a popular classification architecture, which can be used for feature extraction when removing the classification head. It produces 2048-dimensional features. VICReg \cite{bardes2022vicreg} is a feature extraction model trained in a self-supervised manner. For our study, we used a model based on the ResNet50 architecture, which produces 4096-dimensional features. 
Features extracted using models trained in a self-supervised manner (VICReg and DINOv2) seem to adapt better to different downstream tasks, compared to models trained for other specific tasks (ResNet50 on ImageNet), or classical predefined features. The superiority of our proposed model is highlighted both in terms of MAPE and correlation coefficient.

\textbf{Attention Mechanism.} We used two types of attention mechanisms, one attending to the temporal aspect of the data, and the other attending to the individual features. In Table \ref{tab:attention_ablation} we perform an ablation study for three variations for each type of attention mechanism. For the temporal attention, we explored the impact of only using the last frame, multihead attention \cite{vaswani2017attention} (MHA), or the learnable weight vector $\boldsymbol w_t$. Similarly, for the feature-wise attention, we explored the impact of having no such attention, using multihead attention (MHA) or the learnable weight vector $\boldsymbol w_k$. In terms of temporal attention, using $\boldsymbol w_t$ clearly outperforms other methods, especially the multihead attention, while using the last frame still gives reasonable results. $\boldsymbol w_t$ seems to grasp the temporal importance of each frame. 
Concerning the feature-wise attention, the use of multihead attention clearly does not seem adequate for this application, while the effect of incorporating $\boldsymbol w_k$ compared to its absence is not readily apparent. The analysis of its weights could however be used for finer explainability of the model.

\begin{table*}
\begin{center}
\resizebox{0.8\linewidth}{!}{
\begin{tabular}{c c c c c c c c c c}
\toprule
\multicolumn{1}{c}{Temporal attention} & \multicolumn{1}{c}{Last frame} & \multicolumn{1}{c}{Last frame} & \multicolumn{1}{c}{Last frame} & \multicolumn{1}{c}{MHA} & \multicolumn{1}{c}{MHA} & \multicolumn{1}{c}{MHA} & \multicolumn{1}{c}{$\boldsymbol w_t$} & \multicolumn{1}{c}{$\boldsymbol w_t$} & \multicolumn{1}{c}{$\boldsymbol w_t$}  \\
\multicolumn{1}{c}{Feature attention} & \multicolumn{1}{c}{None} & \multicolumn{1}{c}{MHA} & \multicolumn{1}{c}{$\boldsymbol w_k$} & \multicolumn{1}{c}{None} & \multicolumn{1}{c}{MHA} & \multicolumn{1}{c}{$\boldsymbol w_k$} & \multicolumn{1}{c}{None} & \multicolumn{1}{c}{MHA} & \multicolumn{1}{c}{$\boldsymbol w_k$}  \\
\midrule
MAPE $\downarrow$
& 0.2027 & 0.2417 & 0.2070
& 0.2403 & 0.3719 & 0.2382
& \textbf{0.1674} & 0.2320 & \underline{0.1755}\\
Pearson $\uparrow$
& 0.9169 & 0.8358 & 0.9123 
& 0.8755 & 0.6834 & 0.8586
& \underline{0.9209} & 0.8327 & \textbf{0.9214}\\
\bottomrule
\end{tabular}}
\end{center}
\caption{Comparison of MAPE and Pearson correlation coefficient for three types of temporal attention and feature-wise attention. Best results are indicated in \textbf{bold}, and second best results are \underline{underlined}.
}
\label{tab:attention_ablation}
\end{table*}

\textbf{Cavity Aggregation.} In multiple instance learning, the choice of the bag level aggregation plays a crucial role. In Table \ref{tab:aggregation_ablation} we tested multiple methods of aggregation, min-pooling, max-pooling, mean, sum (which is different from the mean as the number of cavities per bag varies), and using a SE \cite{hu2019squeezeandexcitation} block followed by a sum aggregator. The sum and mean operators outperform other types of aggregation, which can be explained by the fact that the ATP is a direct function of the number of organoid cells in the well. This information is not captured by the min-pooling and max-pooling operators. The SE block learns the weight to give to each cavity based on its features, which is used to model the fact that different cavities contribute differently to the ATP. While this may be true, we hypothesize that the information is already present in the features, and the use of the SE block therefore only adds noise.

\begin{table}
\begin{center}
\resizebox{\columnwidth}{!}{
\begin{tabular}{c c c c c c}
\toprule
\multicolumn{1}{c}{Cavity aggregation} & \multicolumn{1}{c}{Min} & \multicolumn{1}{c}{SE~\cite{hu2019squeezeandexcitation}} & \multicolumn{1}{c}{Max} & \multicolumn{1}{c}{Sum} & \multicolumn{1}{c}{Mean}  \\
\midrule
MAPE $\downarrow$
& 0.2425          & 0.2138       
& 0.2092  & \underline{0.1797} & \textbf{0.1755}\\
Pearson $\uparrow$
& 0.8912 & 0.8922
& 0.9185  & \underline{0.9204} & \textbf{0.9214} \\
\bottomrule
\end{tabular}}
\end{center}
\caption{Comparison of MAPE and Pearson correlation coefficient for four methods of feature extraction. Best results are indicated in \textbf{bold}, and second best results are \underline{underlined}.
}
\label{tab:aggregation_ablation}
\vspace{-\baselineskip}
\end{table}

\textbf{\faBullhorn\:Takeaway.} Our model has a mean MAPE $0.1755$ and Pearson of $0.9214$ on the validation set. Trying out different feature extraction methods shows that features from DINOv2 \cite{oquab2023dinov2} provide the best results. Comparing the impact of a variety of feature and temporal attention schemes justifies the use of $\boldsymbol w_t$ and $\boldsymbol w_k$, while we also show the superiority of using the mean as the bag-level aggregation function.


\subsection{Learned Attention Weights}
\label{learned_attention_weights}

\begin{figure}
  \centering
  \begin{subfigure}{.45\linewidth}
    \includegraphics[width=\linewidth]{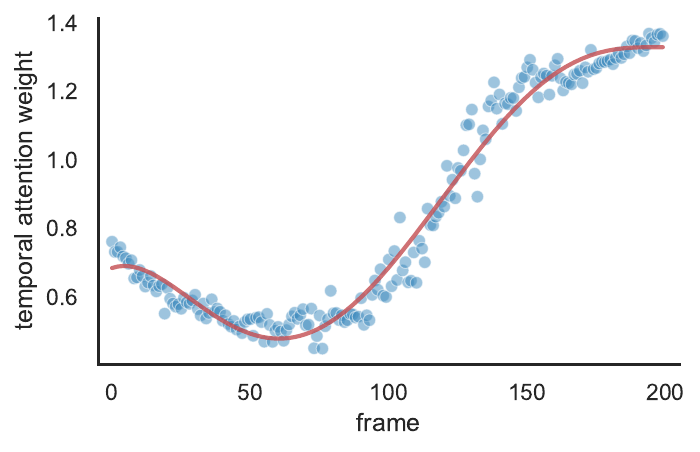}
    \caption{}\label{subfig:temporal_attention}
    \end{subfigure}
    \hfill
    \begin{subfigure}{.45\linewidth}
        \includegraphics[width=\linewidth]{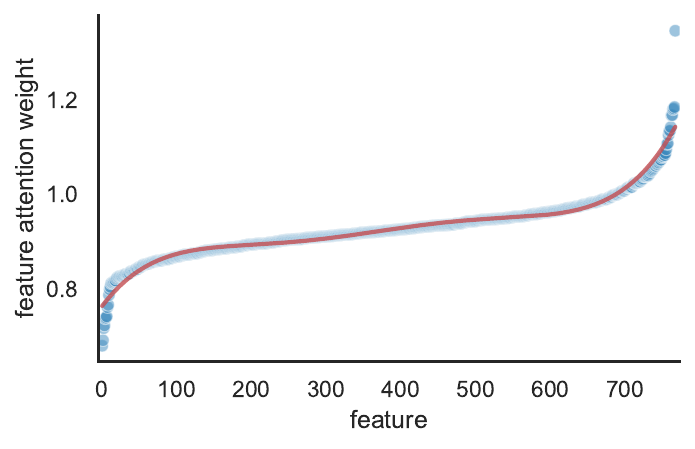}
        \caption{}\label{subfig:feature_attention}
    \end{subfigure}

  \caption{Visualization of the temporal attention (a) and (sorted) feature attention (b) weights for a model after training (visualization for one fold only). Polynomial fit in red.}
\label{fig:visualizing_attention}
\vspace{-\baselineskip}
\end{figure}

Once the model is trained, analyzing the weights learned by the $\boldsymbol w_t$ and $\boldsymbol w_k$ vectors gives insights into our experiments. Figure \ref{subfig:temporal_attention} shows the weight associated with each frame in the $\boldsymbol w_t$ vector after training. Intuitively, we expect later frames to have the most importance. Without any constraints during training, we see that the learned temporal attention has learned a continuous and smooth distribution, meaning that nearby frames will have relatively similar importance. It is noteworthy that the initial frames seem to have more importance for the final prediction than frames after 24 hours of the experiment.
This could be attributed to the fact that the organoid's initial state is significantly determinant of the final ATP, likely due to its size, as larger organoids tend to exhibit higher values of ATP.

Examining the feature importance in Figure \ref{subfig:feature_attention} shows that the feature attention does not highlight any specific feature, as indicated by a ratio of 2 between the feature with the highest and lowest attention weight. Note that in Figure \ref{subfig:feature_attention}, the features are sorted by their attention weight.

\textbf{\faBullhorn\:Takeaway.} The temporal attention vector $\boldsymbol w_t$ learns meaningful relative importance of the time frames, while the feature-wise attention vector $\boldsymbol w_k$ does not make any particular feature stand out.

\subsection{Comparison to SOTA}
To the best of our knowledge, only Ins-ATP \cite{bian2023insatp} have explored the use of machine learning for the prediction of ATP from organoid microscopy images. They imaged multiple organoids placed in a single matrigel drop (compared to our method which has one organoid per cavity) at a single time point, which yields very different visual results to our images. We implemented the Ins-ATP method and adapted it to fit our data, by defining the bags of instances as a set of cavities from the wells, only using the last frame. This can be thought of as a compromise between the presented MeshIns-ATP and DeepIns-ATP \cite{bian2023insatp} as the instances composing the bags are not learned by the model, but are chosen as meaningful regions of interest (the cavities). With this method, we obtained an average MAPE and Pearson correlation coefficient of the 4 validation sets of 0.4375 and 0.6794 respectively (compared to 0.1755 and 0.9214 respectively for our pipeline). This highlights the advances of  our study in the development of a high-throughput screening method to assess real-time drug efficacy from a time-lapse microscopy video of PDOs.

\textbf{\faBullhorn\:Takeaway.} The Ins-ATP \cite{bian2023insatp} method applied to our dataset gives lower performance than our proposed method. 


\subsection{Beyond Predicting Current ATP.}
\label{subsec:beyond_current_atp}
We have shown the ability of our proposed pipeline to, given a sequence of frames, predict the ATP measured at the last frame. Two main questions arise. How well does our model perform in forecasting the ATP? How many frames of history does our model need to predict the current ATP? Note that while we are retraining the model with features from a subset of frames, the segmentation maps are still the ones computed on all $f$ frames.

\begin{figure}[t]
\begin{center}
\begin{subfigure}{.94\linewidth}
    \includegraphics[width=\linewidth]{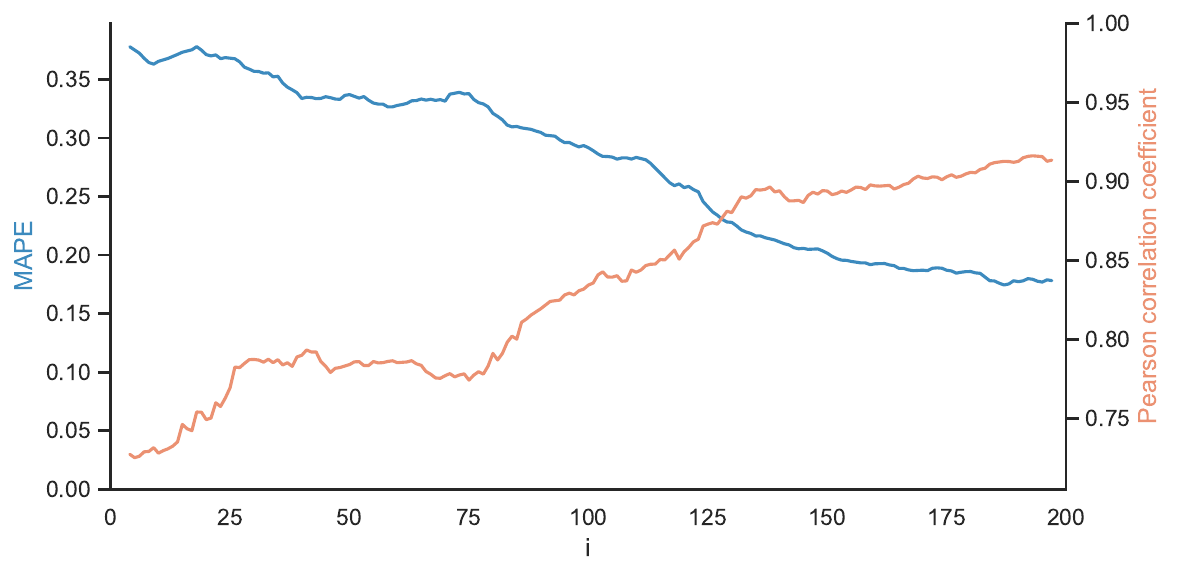}
    \caption{}\label{subfig:hide_last_frames}
\end{subfigure}
\hfill
\begin{subfigure}{.94\linewidth}
    \includegraphics[width=\linewidth]{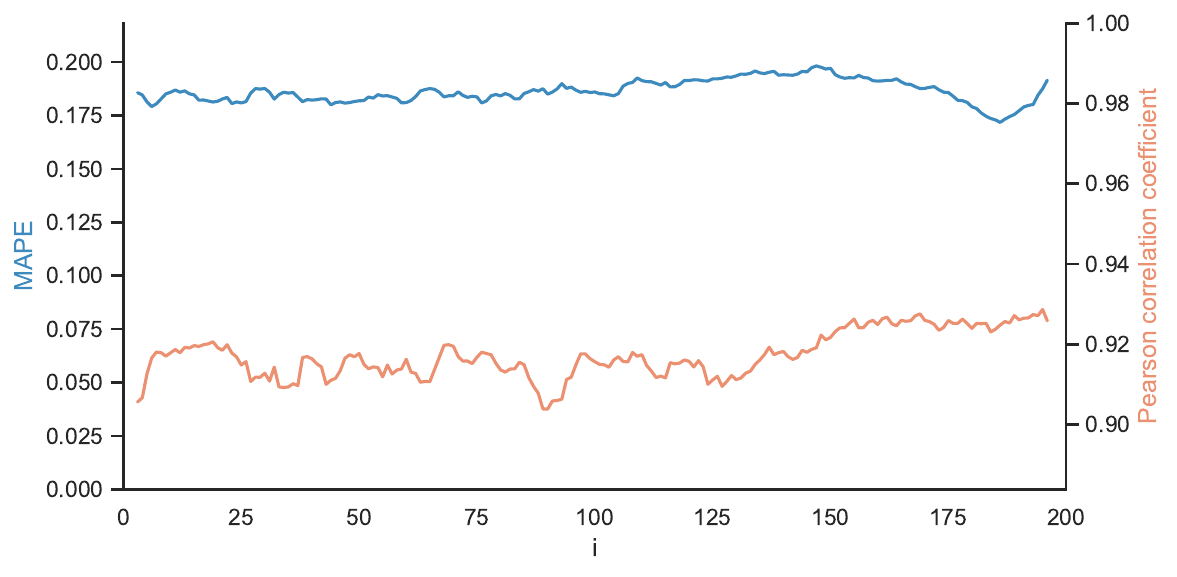}
    \caption{}\label{subfig:show_last_frames}
\end{subfigure}
\vspace{-\baselineskip}
\end{center}
   \caption{MAPE (left) and Pearson correlation coefficient (right) as a function of the number of frames given to the model. (a) Performance of models trained only on frames from 0 to $i$. (b) Performance of models trained only on frames from $i$ to 200.}
\label{fig:hiding_frames}
\vspace{-\baselineskip}
\end{figure}

\textbf{Forecasting the ATP.}
Given our current data (i.e. a sequence of $f=200$ frames, and a measure of ATP at the last frame), we can evaluate the performance of our pipeline for forecasting the ATP by training the model while omitting the last frames. In Figure \ref{subfig:hide_last_frames}, we show the performance of the model when trained only on frames from 0 to $i$, equivalent to predicting the ATP $(f-i)$ frames in advance. As expected, the performance of the model drops as the model is trained to predict ATPs earlier in the video. However, predicting 2 days in advance a MAPE of $0.28$ seems reasonable compared to the best performance of $0.1755$, especially when considering the wide range of ATP values.

\textbf{Required history for ATP prediction.}
Similarly, if we omit the first frames from the training, we can evaluate how many frames of history our model needs for predicting the ATP. In Figure \ref{subfig:show_last_frames} we show the performance of the model when trained only on frames from $i$ to $f$ (i.e. having access to $f-i$ frames). The drop in performance is not as clear as the one shown in Figure \ref{subfig:hide_last_frames}. This indicates that our model seems to perform well when not taking too many frames into account: the peak performance appears to happen when about 15 frames are given to the model. This is positive because it suggests that our model could be used for the live measurement of ATP as the images are still being acquired, as it does not need a full history of the organoid cells. However, it is important to note that although the ATP prediction part of the model seems to perform well with only 15 frames, the segmentation pipeline (and more precisely, the generation of the first prompt for SAM) relies on movement within the timelapse. We tested how well the segmentation of the organoids worked on the last 15 frames by computing the Dice score with the last 15 frames of the segmentation map computed using all frames as ground truth. 
The mean Dice score is 0.80, with a very uneven distribution among cavities: 75\% of the cavities obtain a Dice over 0.90, while 15\% obtain a Dice under 0.20. It seems that most segmentations are not affected by the use of a smaller number of frames, but for those cases which are affected, the segmentations are prone to complete failure.

\textbf{\faBullhorn\:Takeaway.} Our model can predict the ATP in advance with a reasonable drop in performance and is able to predict with only about 15 frames of history making it usable for online predictions.

\section{Conclusion}
The estimation of the ATP is a standard method for the estimation of drug efficacy on organoids. However, it allows the measurement at a single timepoint for the entire experiment. In this paper, we propose a method for the spatio-temporal analysis of organoid microscopy timelapse videos, for the prediction of ATP. We assess the performance of our approach for predicting the current ATP with different ablations and we report better performance than SOTA.

Future work includes the further exploration of foundation models for the analysis of organoid videos. In this study, the foundation models are used as frozen blocks. However, authors of DINOv2 \cite{oquab2023dinov2} showed that finetuning the model encoders to a specific dataset improved the results on the dataset. One step could be to finetune the feature extraction model to improve the representations of the organoids to be better adapted to the prediction of ATP. Similarly for the SAM \cite{kirillov2023segany} model, we could finetune the mask decoder part of the model, or learn more efficient prompts.

\textbf{Acknowledgments.}
This work has benefited from state financial aid, managed by the Agence Nationale de Recherche under the investment program integrated into France 2030, project reference ANR-21-RHUS-0003. Experiments have been conducted using HPC resources from the \href{http://mesocentre.centralesupelec.fr/}{“Mésocentre”} computing center.

{\small
\bibliographystyle{ieee_fullname}
\bibliography{main}

\begin{thebibliography}{10}\itemsep=-1pt

\bibitem{Abdul2022DCryptO}
Lyan Abdul, Jocelyn Xu, Alexander Sotra, Abbas Chaudary, Jerry Gao, Shravanthi
  Rajasekar, Nicky Anvari, Hamidreza Mahyar, and Boyang Zhang.
\newblock D-crypto: deep learning-based analysis of colon organoid morphology
  from brightfield images.
\newblock {\em Lab Chip}, 22:4118--4128, 2022.

\bibitem{adams2022does}
Lisa~C. Adams, Felix Busch, Daniel Truhn, Marcus~R. Makowski, Hugo~JWL. Aerts,
  and Keno~K. Bressem.
\newblock What does dall-e 2 know about radiology?, 2022.

\bibitem{arnab2021vivit}
Anurag Arnab, Mostafa Dehghani, Georg Heigold, Chen Sun, Mario Lučić, and
  Cordelia Schmid.
\newblock Vivit: A video vision transformer, 2021.

\bibitem{bardes2022vicreg}
Adrien Bardes, Jean Ponce, and Yann LeCun.
\newblock Vicreg: Variance-invariance-covariance regularization for
  self-supervised learning.
\newblock In {\em ICLR}, 2022.

\bibitem{bertasius2021spacetime}
Gedas Bertasius, Heng Wang, and Lorenzo Torresani.
\newblock Is space-time attention all you need for video understanding?, 2021.

\bibitem{bian2021tracking}
Xuesheng Bian, Gang Li, Cheng Wang, Weiquan Liu, Xiuhong Lin, Zexin Chen,
  Mancheung Cheung, and Xiongbiao Luo.
\newblock A deep learning model for detection and tracking in high-throughput
  images of organoid.
\newblock {\em Computers in Biology and Medicine}, 134:104490, 2021.

\bibitem{bian2023insatp}
Xuesheng Bian, Cheng Wang, Shuting Chen, Weiquan Liu, Sen Xu, Jinxin Zhu,
  Rugang Wang, Zexin Chen, Min Huang, and Gang Li.
\newblock Ins-atp: Deep estimation of atp for organoid based on high throughput
  microscopic images, 2023.

\bibitem{canny1986computational}
John Canny.
\newblock A computational approach to edge detection.
\newblock {\em IEEE Transactions on pattern analysis and machine intelligence},
  (6):679--698, 1986.

\bibitem{caron2021emerging}
Mathilde Caron, Hugo Touvron, Ishan Misra, Hervé Jégou, Julien Mairal, Piotr
  Bojanowski, and Armand Joulin.
\newblock Emerging properties in self-supervised vision transformers, 2021.

\bibitem{chen2023videollm}
Guo Chen, Yin-Dong Zheng, Jiahao Wang, Jilan Xu, Yifei Huang, Junting Pan, Yi
  Wang, Yali Wang, Yu Qiao, Tong Lu, and Limin Wang.
\newblock Videollm: Modeling video sequence with large language models, 2023.

\bibitem{chen2020simple}
Ting Chen, Simon Kornblith, Mohammad Norouzi, and Geoffrey Hinton.
\newblock A simple framework for contrastive learning of visual
  representations.
\newblock {\em arXiv preprint arXiv:2002.05709}, 2020.

\bibitem{deng2009imagenet}
Jia Deng, Wei Dong, Richard Socher, Li-Jia Li, Kai Li, and Li Fei-Fei.
\newblock Imagenet: A large-scale hierarchical image database.
\newblock In {\em 2009 IEEE conference on computer vision and pattern
  recognition}, pages 248--255. Ieee, 2009.

\bibitem{dosovitskiy2021image}
Alexey Dosovitskiy, Lucas Beyer, Alexander Kolesnikov, Dirk Weissenborn,
  Xiaohua Zhai, Thomas Unterthiner, Mostafa Dehghani, Matthias Minderer, Georg
  Heigold, Sylvain Gelly, Jakob Uszkoreit, and Neil Houlsby.
\newblock An image is worth 16x16 words: Transformers for image recognition at
  scale, 2021.

\bibitem{goyal2021selfsupervised}
Priya Goyal, Mathilde Caron, Benjamin Lefaudeux, Min Xu, Pengchao Wang, Vivek
  Pai, Mannat Singh, Vitaliy Liptchinsky, Ishan Misra, Armand Joulin, and Piotr
  Bojanowski.
\newblock Self-supervised pretraining of visual features in the wild, 2021.

\bibitem{gu2022efficiently}
Albert Gu, Karan Goel, and Christopher Ré.
\newblock Efficiently modeling long sequences with structured state spaces,
  2022.

\bibitem{he2015resnet}
Kaiming He, Xiangyu Zhang, Shaoqing Ren, and Jian Sun.
\newblock Deep residual learning for image recognition.
\newblock {\em CoRR}, abs/1512.03385, 2015.

\bibitem{he2015prelu}
Kaiming He, Xiangyu Zhang, Shaoqing Ren, and Jian Sun.
\newblock Delving deep into rectifiers: Surpassing human-level performance on
  imagenet classification, 2015.

\bibitem{hochreiter1997long}
Sepp Hochreiter and J{\"u}rgen Schmidhuber.
\newblock Long short-term memory.
\newblock {\em Neural computation}, 9(8):1735--1780, 1997.

\bibitem{hu2019squeezeandexcitation}
Jie Hu, Li Shen, Samuel Albanie, Gang Sun, and Enhua Wu.
\newblock Squeeze-and-excitation networks, 2019.

\bibitem{kanopoulos1988design}
Nick Kanopoulos, Nagesh Vasanthavada, and Robert~L Baker.
\newblock Design of an image edge detection filter using the sobel operator.
\newblock {\em IEEE Journal of solid-state circuits}, 23(2):358--367, 1988.

\bibitem{kather2022medical}
Jakob~Nikolas Kather, Narmin~Ghaffari Laleh, Sebastian Foersch, and Daniel
  Truhn.
\newblock Medical domain knowledge in domain-agnostic generative {AI}.
\newblock {\em npj Digital Medicine}, 5(1), July 2022.

\bibitem{kirillov2023segany}
Alexander Kirillov, Eric Mintun, Nikhila Ravi, Hanzi Mao, Chloe Rolland, Laura
  Gustafson, Tete Xiao, Spencer Whitehead, Alexander~C. Berg, Wan-Yen Lo, Piotr
  Doll{\'a}r, and Ross Girshick.
\newblock Segment anything.
\newblock {\em arXiv:2304.02643}, 2023.

\bibitem{letai2017functional}
Anthony Letai.
\newblock Functional precision cancer medicine{\textemdash}moving beyond pure
  genomics.
\newblock {\em Nature Medicine}, 23(9):1028--1035, Sept. 2017.

\bibitem{li2022blip}
Junnan Li, Dongxu Li, Caiming Xiong, and Steven Hoi.
\newblock Blip: Bootstrapping language-image pre-training for unified
  vision-language understanding and generation, 2022.

\bibitem{li2022grounded}
Liunian~Harold Li, Pengchuan Zhang, Haotian Zhang, Jianwei Yang, Chunyuan Li,
  Yiwu Zhong, Lijuan Wang, Lu Yuan, Lei Zhang, Jenq-Neng Hwang, Kai-Wei Chang,
  and Jianfeng Gao.
\newblock Grounded language-image pre-training, 2022.

\bibitem{liu2023samm}
Yihao Liu, Jiaming Zhang, Zhangcong She, Amir Kheradmand, and Mehran Armand.
\newblock Samm (segment any medical model): A 3d slicer integration to sam,
  2023.

\bibitem{liu2021video}
Ze Liu, Jia Ning, Yue Cao, Yixuan Wei, Zheng Zhang, Stephen Lin, and Han Hu.
\newblock Video swin transformer, 2021.

\bibitem{loshchilov2019adamw}
Ilya Loshchilov and Frank Hutter.
\newblock Decoupled weight decay regularization, 2019.

\bibitem{Lowe2004}
David~G. Lowe.
\newblock Distinctive image features from scale-invariant keypoints.
\newblock {\em International Journal of Computer Vision}, 60(2):91--110, Nov.
  2004.

\bibitem{Matthews2022OrganoID}
Jonathan Matthews, Brooke Schuster, Sara~Saheb Kashaf, Ping Liu, Mustafa
  Bilgic, Andrey Rzhetsky, and Sava{\c s} Tay.
\newblock Organoid: a versatile deep learning platform for organoid image
  analysis.
\newblock {\em bioRxiv}, 2022.

\bibitem{Metzger2022Huntington}
Jakob~J. Metzger, Carlota Pereda, Arjun Adhikari, Tomomi Haremaki, Szilvia
  Galgoczi, Eric~D. Siggia, Ali~H. Brivanlou, and Fred Etoc.
\newblock Deep-learning analysis of micropattern-based organoids enables
  high-throughput drug screening of huntington’s disease models.
\newblock {\em Cell Reports Methods}, 2(9):100297, 2022.

\bibitem{mu2021slip}
Norman Mu, Alexander Kirillov, David Wagner, and Saining Xie.
\newblock Slip: Self-supervision meets language-image pre-training, 2021.

\bibitem{nichol2022glide}
Alex Nichol, Prafulla Dhariwal, Aditya Ramesh, Pranav Shyam, Pamela Mishkin,
  Bob McGrew, Ilya Sutskever, and Mark Chen.
\newblock Glide: Towards photorealistic image generation and editing with
  text-guided diffusion models, 2022.

\bibitem{oquab2023dinov2}
Maxime Oquab, Timothée Darcet, Theo Moutakanni, Huy~V. Vo, Marc Szafraniec,
  Vasil Khalidov, Pierre Fernandez, Daniel Haziza, Francisco Massa, Alaaeldin
  El-Nouby, Russell Howes, Po-Yao Huang, Hu Xu, Vasu Sharma, Shang-Wen Li,
  Wojciech Galuba, Mike Rabbat, Mido Assran, Nicolas Ballas, Gabriel Synnaeve,
  Ishan Misra, Herve Jegou, Julien Mairal, Patrick Labatut, Armand Joulin, and
  Piotr Bojanowski.
\newblock Dinov2: Learning robust visual features without supervision, 2023.

\bibitem{Park2023Kidney}
Keonhyeok Park, Jong~Young Lee, Soo~Young Lee, Iljoo Jeong, Seo-Yeon Park,
  Jin~Won Kim, Sun~Ah Nam, Hyung~Wook Kim, Yong~Kyun Kim, and Seungchul Lee.
\newblock Deep learning predicts the differentiation of kidney organoids
  derived from human induced pluripotent stem cells.
\newblock {\em Kidney Research and Clinical Practice}, 42(1):75--85, Jan. 2023.

\bibitem{paszke2019pytorch}
Adam Paszke, Sam Gross, Francisco Massa, Adam Lerer, James Bradbury, Gregory
  Chanan, Trevor Killeen, Zeming Lin, Natalia Gimelshein, Luca Antiga, Alban
  Desmaison, Andreas Kopf, Edward Yang, Zachary DeVito, Martin Raison, Alykhan
  Tejani, Sasank Chilamkurthy, Benoit Steiner, Lu Fang, Junjie Bai, and Soumith
  Chintala.
\newblock Pytorch: An imperative style, high-performance deep learning library.
\newblock In H. Wallach, H. Larochelle, A. Beygelzimer, F. d\textquotesingle
  Alch\'{e}-Buc, E. Fox, and R. Garnett, editors, {\em Advances in Neural
  Information Processing Systems}, volume~32. Curran Associates, Inc., 2019.

\bibitem{phan2019simple}
Nhan Phan, Jenny~J. Hong, Bobby Tofig, Matthew Mapua, David Elashoff, Neda~A.
  Moatamed, Jin Huang, Sanaz Memarzadeh, Robert Damoiseaux, and Alice Soragni.
\newblock A simple high-throughput approach identifies actionable drug
  sensitivities in patient-derived tumor organoids.
\newblock {\em Communications Biology}, 2(1), Feb. 2019.

\bibitem{radford2021learning}
Alec Radford, Jong~Wook Kim, Chris Hallacy, Aditya Ramesh, Gabriel Goh,
  Sandhini Agarwal, Girish Sastry, Amanda Askell, Pamela Mishkin, Jack Clark,
  Gretchen Krueger, and Ilya Sutskever.
\newblock Learning transferable visual models from natural language
  supervision, 2021.

\bibitem{ramesh2021zeroshot}
Aditya Ramesh, Mikhail Pavlov, Gabriel Goh, Scott Gray, Chelsea Voss, Alec
  Radford, Mark Chen, and Ilya Sutskever.
\newblock Zero-shot text-to-image generation, 2021.

\bibitem{schneider2012image}
Caroline~A Schneider, Wayne~S Rasband, and Kevin~W Eliceiri.
\newblock Nih image to imagej: 25 years of image analysis.
\newblock {\em Nat Meth}, 9(7):671--675, July 2012.

\bibitem{vangriethuysen2017computational}
Joost~J.M. van Griethuysen, Andriy Fedorov, Chintan Parmar, Ahmed Hosny, Nicole
  Aucoin, Vivek Narayan, Regina~G.H. Beets-Tan, Jean-Christophe Fillion-Robin,
  Steve Pieper, and Hugo~J.W.L. Aerts.
\newblock Computational radiomics system to decode the radiographic phenotype.
\newblock {\em Cancer Research}, 77(21):e104--e107, Oct. 2017.

\bibitem{vaswani2017attention}
Ashish Vaswani, Noam Shazeer, Niki Parmar, Jakob Uszkoreit, Llion Jones,
  Aidan~N Gomez, \L~ukasz Kaiser, and Illia Polosukhin.
\newblock Attention is all you need.
\newblock In I. Guyon, U.~Von Luxburg, S. Bengio, H. Wallach, R. Fergus, S.
  Vishwanathan, and R. Garnett, editors, {\em Advances in Neural Information
  Processing Systems}, volume~30. Curran Associates, Inc., 2017.

\bibitem{wang2023seggpt}
Xinlong Wang, Xiaosong Zhang, Yue Cao, Wen Wang, Chunhua Shen, and Tiejun
  Huang.
\newblock Seggpt: Segmenting everything in context, 2023.

\bibitem{wu2023medicalsam}
Junde Wu, Yu Zhang, Rao Fu, Huihui Fang, Yuanpei Liu, Zhaowei Wang, Yanwu Xu,
  and Yueming Jin.
\newblock Medical sam adapter: Adapting segment anything model for medical
  image segmentation, 2023.

\bibitem{yang2023track}
Jinyu Yang, Mingqi Gao, Zhe Li, Shang Gao, Fangjing Wang, and Feng Zheng.
\newblock Track anything: Segment anything meets videos, 2023.

\bibitem{yuan2021florence}
Lu Yuan, Dongdong Chen, Yi-Ling Chen, Noel Codella, Xiyang Dai, Jianfeng Gao,
  Houdong Hu, Xuedong Huang, Boxin Li, Chunyuan Li, Ce Liu, Mengchen Liu,
  Zicheng Liu, Yumao Lu, Yu Shi, Lijuan Wang, Jianfeng Wang, Bin Xiao, Zhen
  Xiao, Jianwei Yang, Michael Zeng, Luowei Zhou, and Pengchuan Zhang.
\newblock Florence: A new foundation model for computer vision, 2021.

\bibitem{zerveas2020transformerbased}
George Zerveas, Srideepika Jayaraman, Dhaval Patel, Anuradha Bhamidipaty, and
  Carsten Eickhoff.
\newblock A transformer-based framework for multivariate time series
  representation learning, 2020.

\bibitem{zhang2023effectively}
Michael Zhang, Khaled~K. Saab, Michael Poli, Tri Dao, Karan Goel, and
  Christopher Ré.
\newblock Effectively modeling time series with simple discrete state spaces,
  2023.

\bibitem{zhao2023fast}
Xu Zhao, Wenchao Ding, Yongqi An, Yinglong Du, Tao Yu, Min Li, Ming Tang, and
  Jinqiao Wang.
\newblock Fast segment anything, 2023.

\bibitem{zou2023segment}
Xueyan Zou, Jianwei Yang, Hao Zhang, Feng Li, Linjie Li, Jianfeng Gao, and
  Yong~Jae Lee.
\newblock Segment everything everywhere all at once, 2023.

\end{thebibliography}
}

\end{document}